\documentclass[10pt,twocolumn,letterpaper]{article}

\usepackage[pagenumbers]{cvpr} 

\definecolor{cvprblue}{rgb}{0.21,0.49,0.74}
\usepackage[pagebackref,breaklinks,colorlinks,allcolors=cvprblue]{hyperref}
\usepackage{multirow}
\usepackage{pifont}
\usepackage{booktabs}
\usepackage{colortbl}

\def\paperID{11517} 
\def\confName{CVPR}
\def\confYear{2026}

\title{DeCo-VAE: Learning Compact Latents for Video Reconstruction via Decoupled Representation}

\author{
Xiangchen Yin$^{1}$ \quad Jiahui Yuan$^{1}$ \quad Zhangchi Hu$^{1}$ \quad Wenzhang Sun$^{2}$\thanks{Project leader.} \\  Jie Chen$^{1}$ \quad Xiaozhen Qiao$^{1}$ \quad Hao Li$^{2}$ 
\quad Xiaoyan Sun$^{1}$\thanks{Corresponding author.} \vspace{6pt} \\ 
$^1$University of Science and Technology of China \\
$^2$Li Auto Inc. \\
{\tt\small  yinxiangchen@mail.ustc.edu.cn 
}
}

\begin{document}
\maketitle
\begin{abstract}

Existing video Variational Autoencoders (VAEs) generally overlook the similarity between frame contents, leading to redundant latent modeling. In this paper, we propose decoupled VAE (DeCo-VAE) to achieve compact latent representation. Instead of encoding RGB pixels directly, we decompose video content into distinct components via explicit decoupling: keyframe, motion and residual, and learn dedicated latent representation for each. To avoid cross-component interference, we design dedicated encoders for each decoupled component and adopt a shared 3D decoder to maintain spatiotemporal consistency during reconstruction. We further utilize a decoupled adaptation strategy that freezes partial encoders while training the others sequentially, ensuring stable training and accurate learning of both static and dynamic features. Extensive quantitative and qualitative experiments demonstrate that DeCo-VAE achieves superior video reconstruction performance.

\end{abstract}    
\section{Introduction}
\label{sec:intro}

Video Variational Autoencoders (VAEs) transforms video frames into compact latent representation as a critical component of Latent Video Diffusion Models (LVDMs)~\cite{ldm, videocrafter2}. Several models such as Sora~\cite{sora}, Open-Sora-Plan~\cite{open_sora_plan}, CogVideoX~\cite{cogvideox}, Stable Video Diffusion~\cite{svd} have achieved powerful performance, the efficiency and quality of this process directly impact the performance of downstream generation tasks.

Early video generation methods directly adopted image VAEs~\cite{ldm} in latent representation to perform video compression through frame-by-frame encoding. These methods fail to capture the temporal correlations between frames, essentially reducing videos to sequences of independent images. To solve this limitation, several methods~\cite{cogvideox, odvae, open_sora_plan} have employed dense 3D networks with heavy parameters to enhance spatiotemporal interactions. While improving reconstruction quality, these approaches cause exponential growth in network parameters and computational complexity, significantly compromising video reconstruction efficiency. In contrast, other methods~\cite{sora} use lightweight 2+1D architectures, reducing computational costs through separating spatial and temporal convolutions. However, such lightweight designs struggle to model complex video dynamics and temporal dependencies. To balance efficiency and quality, recent advances in video VAEs~\cite{wfvae,reduciovae,videovaeplus,liu2025hi} have leveraged lightweight designs such as wavelet transforms, reducing computational overhead while better preserving critical visual information. Additionally, some approaches~\cite{vidtwin} establish different latent spaces to capture dynamics, but still cannot effectively decouple the motion information of the video.

\begin{figure*}[t]
\begin{center}

\includegraphics[width=0.98\textwidth]{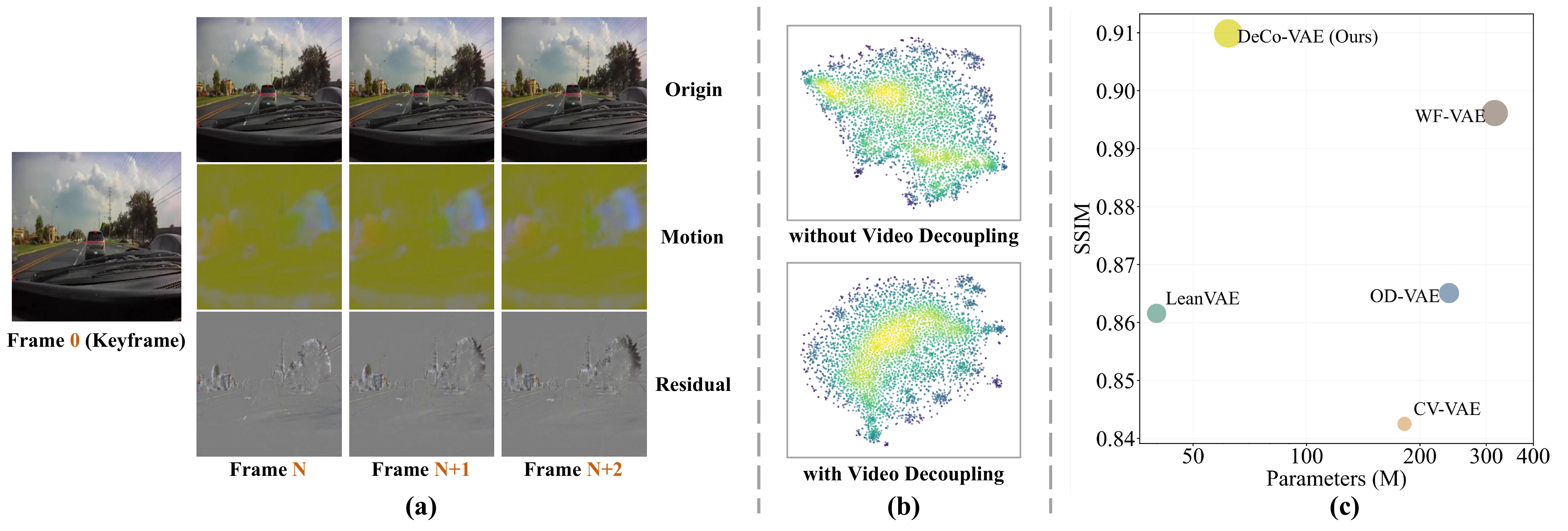}

\end{center}
\caption{(a) Visualization of decoupled components in DeCo-VAE, including keyframe, motion and residual components for video frames. (b) Visualization of t-SNE latent distributions in video decoupling, our DeCo-VAE achieves more compact latent space. (c) Performance comparison of video VAEs, our DeCo-VAE achieves superior reconstruction quality with lightweight parameters.} \label{Intro}
\end{figure*}

Despite some progress made by these methods, they treats videos as monolithic data, without considering the high redundancy between consecutive frames. This creates a paradox: while video data is inherently highly redundant and should be easier to compress, lightweight architectures struggle to effectively leverage this redundancy for simplified modeling. Conversely, heavy networks capable of comprehensive modeling introduce unnecessary computational overhead for handling such redundant content. Video Codec~\cite{mpeg,hevc,compressai} decomposes videos into keyframe, motion and residual components, as shown in Fig.~\ref{Intro} (a), keyframe contains static texture information and spatial structures, while residual and motion components only represent temporal differences. This line of thinking effectively removes redundancy in videos, offering a new research perspective for current video VAEs methods. By visualizing the latent distributions (both with and without video decoupling, as shown in Fig.~\ref{Intro} (b)), we observe that the latent space learned via video decoupling is significantly more compact, demonstrates the effectiveness of video de-redundancy. This tighter concentration of the latent distribution is validated by the clustered highlight regions in the visualization results.

In this paper, we propose decoupled VAE (DeCo-VAE), a video VAE framework utilizing explicit video content decoupling. Instead of directly encoding raw pixels, we construct latent representation for the decoupled motion, residual and keyframe. The motion component focuses on inter-frame dynamic differences and the residual component captures fine-grained details, while the keyframe serve as appearance anchors to preserve basic textures and spatial structures. To avoid cross-interference between these components, we assign dedicated encoders to eliminate feature entanglement and use a shared 3D decoder to restore three latent representation, maintaining spatiotemporal consistency during reconstruction. To enhance training stability under complex constraint conditions, We employ a decoupled adaptation strategy, different encoders are frozen in each phase to train the counterpart sequentially. This staged approach avoids cross-component feature interference, ensuring precise learning of both static and dynamic features. Consequently, Our decoupled design enables strong performance on lightweight VAEs (Fig.~\ref{Intro} (c)), its latent representation serve as an efficient drop-in solution for downstream video generation tasks.

Our main contributions are summarized as follows:
\begin{itemize}
\item We propose DeCo-VAE, a lightweight video VAE framework for explicit decoupled modeling of video content. By decomposing motion dynamics and fine-grained residual details with keyframes as appearance anchors, it avoids feature entanglement in raw pixel encoding, enhancing reconstruction quality and representation interpretability.

\item We integrate dedicated encoders for each decoupled component and a shared 3D decoder to maintain spatiotemporal consistency. Alongside a decoupled adaptation strategy that freezes partial encoder, which eliminates cross-component interference and ensures stable training with precise learning of static and dynamic features.

\item Leveraging its decoupled design, DeCo-VAE enables superior performance on video reconstruction, whose latent representation serve as an efficient drop-in solution for downstream video generation tasks.
\end{itemize}

\section{Related Work}
\label{sec:relat}

\subsection{Video Diffusion Models}

Latent Video Diffusion Models (LVDMs) have emerged as the cornerstone of state-of-the-art video generation, powering flagship frameworks such as Sora~\cite{sora}, OpenSora~\cite{zheng2024open}, Open Sora Plan~\cite{open_sora_plan}, VideoCrafter~\cite{chen2023videocrafter1, videocrafter2}, Latte~\cite{latte}, CogVideoX~\cite{cogvideox}, DynamiCrafter~\cite{xing2024dynamicrafter}, Vidu~\cite{bao2024vidu}, and Hunyuan Video~\cite{kong2024hunyuanvideo}. Beyond general video synthesis, LVDMs also enable specialized tasks including controllable video generation~\cite{he2023animate} and multimodal video generation~\cite{he2024llms}.

The LVDMs pipeline follows a two-stage paradigm: first, a video Variational Autoencoders (VAE) compresses raw video data into a compact latent space, drastically reducing computational costs; second, a noise prediction model operates within this latent domain to learn and perform target transformations. The performance of LVDMs is inherently tied to the quality of the video VAE, as generated video fidelity depends critically on both the representational capacity of the latent space and the VAE’s encoding-decoding efficiency.

In image generation, frameworks like the Stable Diffusion series~\cite{podell2023sdxl, rombach2022high, grpose} have achieved remarkable success, largely due to their efficient VAEs that enable high-fidelity reconstruction across diverse image types. By contrast, existing video VAEs have not matched this performance. This gap arises from the unique challenges of compressing video with spatiotemporal correlations while maintaining compactness remains an unresolved hurdle. Consequently, LVDMs are often constrained in scenarios with complex motion, limiting their ability to generate temporally coherent, high-quality videos.

\subsection{Video Variational Autoencoder}

Video Variational Autoencoders (VAEs) are key for latent video compression in Latent Video Diffusion Models (LVDMs), divided into discrete and continuous types. Discrete VAEs like MAGVIT-v2~\cite{yu2023language} enable high-quality reconstruction but lack backpropagation gradients, making them incompatible with LVDMs. Continuous VAEs (e.g., Stable Video Diffusion~\cite{blattmann2023stable}) are widely used in LVDMs, evolving $4 \times 8 \times 8$ compression to reduce temporal redundancy, yet most struggle with large-motion video reconstruction due to weak temporal modeling.

To balance efficiency and spatiotemporal performance, existing methods take diverse approaches: dense 3D networks (e.g., OD-VAE~\cite{chen2024od}) boost interactions but increase computation. lightweight 2D+1D architectures (e.g., Open-Sora~\cite{zheng2024open}) cut costs, but cause motion blurring. WF-VAE~\cite{li2025wf} adopts multi-level wavelet transform to leverage low frequency energy flow for latent representation and design causal cache to achieve block-wise prediction for long video reconstruction. VidTwin~\cite{vidtwin} encode  distinct latent spaces to respresents the structure vector and  dynamics latent vectors. OmniTokenizer~\cite{wang2024omnitokenizer} adopts a space-time decoupling architecture design, integrating windows and causal attention for space-time modeling, but they cannot effectively decouple the motion features and static features in the video. LeanVAE~\cite{cheng2025leanvae} integrates wavelet transforms and compressed sensing to balance efficiency and reconstruction quality, and supports LVDMs by addressing high-resolution or large-motion video compression bottlenecks, but increasing its latent channel count fails to improve generation performance and even causes video distortion. Notably, these methods fail to leverage interframe similarities, limiting content-aware representation.

In decoupled modeling, inspired by codecs like MPEG-4 (e.g., Video-LaViT~\cite{jin2024video}), which decompose videos into keyframes and motion but either target specific video types or lack full decoupling, missing fine-grained residual modeling. Thus, a video VAE that explicitly decouples content, avoids cross-interference, and stabilizes static-dynamic feature learning is needed for better large-motion reconstruction.

\subsection{Decoupled Video Models}


Video compression remains a fundamental challenge in computer vision. Recent approaches have adopted a disentangled paradigm: traditional codecs like MPEG-4~\cite{legall1993video} use I-frames for keyframe representation and motion vectors to capture dynamics. Inspired by this, Video-LaViT encodes~\cite{jin2024video} keyframes and motion vectors into tokens for integration with large language models. Other representative motion representation include MotionI2V~\cite{shi2024motion}, which models pixel trajectories, and methods leveraging optical flow~\cite{lew2025disentangled} for frame interpolation. Some works target specific video types, the GAIA series~\cite{he2023gaia,wang2025instructavatar,yu2024make} focuses on talking faces by disentangling identity and motion via self-cross reenactment, while iVideoGPT~\cite{wu2024ivideogpt} explores embodied video modeling. D-VDM~\cite{shen2024decouple} designs diffusion-based models that explicitly disentangle spatial content including object shapes, texture layouts, motion vectors encoding inter-frame geometric transformations and residual components capturing fine-grained details unaccounted for by motion warping, aiming to address the inefficiencies and temporal inconsistency issues in conditional image-to-video generation caused by feature entanglement between static and dynamic information in RGB pixel space. CMD~\cite{yu2024efficient} represents content via a weighted average of all frames serving as the common content encoded by an autoencoder and models motion as a low-dimensional latent representation, which is learned by a new lightweight diffusion model to enable efficient video generation while leveraging a pretrained image diffusion model for improved quality.

In contrast, our method does not encode raw pixels directly. Instead, we learn disentangled latent representation for video VAE: the motion branch models inter-frame dynamics, the residual branch captures fine details, and a keyframe serves as an appearance anchor to preserve texture and spatial structure.

\section{DeCo-VAE Approach}

\begin{figure*}[t]
\begin{center}

\includegraphics[width=1\textwidth]{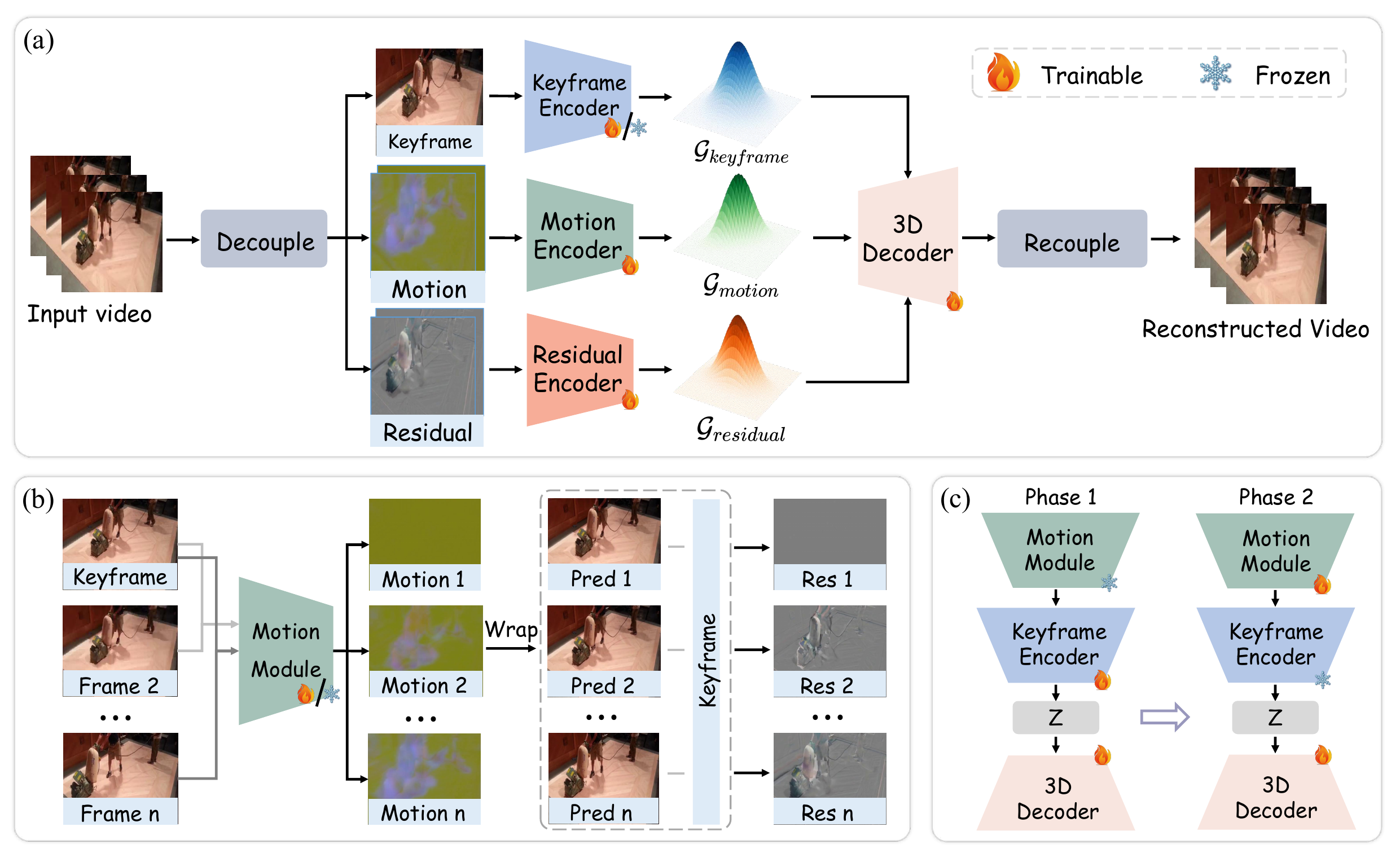}

\end{center}
\caption{\textbf{Overview of the proposed DeCo-VAE.} (a) DeCo-VAE pipeline decomposes video sequences into keyframe, motion, and residual, via dedicated encoders and a shared 3D decoder. (b) With the keyframe as reference, subsequent frames (with keyframe) are inputs to a motion module for motion components, motion compensation generates predicted frames, and residuals are obtained by subtracting predicted frames from keyframe. (c) Decoupled adaptation strategy stabilizes training and enhances temporal consistency. } \label{Overview}
\end{figure*}

\subsection{Overall Architecture}

We propose the DeCo-VAE framework with the overall architecture illustrated in Fig.~\ref{Overview} (a), aiming to learn precise latent representation by explicitly decoupling video content into semantically distinct components. Unlike previous video VAEs that directly encode raw pixels, DeCo-VAE decomposes videos into three mutually exclusive components: keyframes, motion, and residuals. The framework adopts three dedicated encoders for decoupled components and a shared 3D decoder to avoid cross-interference. Moreover, we design a decoupled adaptation strategy freezing one encoder while training the other sequentially to ensure stable training and precise component-specific feature learning.

We decompose an original video $\mathcal{X}$ into motion $\mathcal{X}_{\text{m}}$ and residual $\mathcal{X}_{\text{r}}$, while we select the first frame of the input as the keyframe $\mathcal{X}_{\text{k}}$, preserving critical information in the whole video sequence. Specifically, the motion component with dynamic geometric transformations is extracted via a pre-trained motion module $\mathcal{M}$~\cite{agustsson2020scale}, and the residual component focuses on details not covered by motion prediction and preserves fine-grained features. This decoupling process significantly reduces redundant information in the original video. After receiving the decoder outputs from the video VAE, we restore the reconstructed video through the recoupling operation, ensuring the spatiotemporal consistency and fidelity of the reconstructed results.

Equipped with a 3D encoder-decoder architecture, video VAE learns compact latent representation for the decoupled components. The encoder module consists of three dedicated encoders ($\mathcal{E}_{\text{k}}$, $\mathcal{E}_{\text{m}}$, $\mathcal{E}_{\text{r}}$), each with downsampling layers and residual blocks to capture spatiotemporal features for keyframe, motion, and residual, respectively. These encoders map their inputs to parameters of latent Gaussian distributions:
\begin{equation}
\mu_i, \log\sigma_i^2 = \mathcal{E}_i(\mathcal{X}_i), \quad i \in \{\text{k}, \text{m}, \text{r}\}
\end{equation}
where $\mu_{i}, \log\sigma_{i}^2 \in \mathbb{R}^{D \times T' \times H' \times W'}$, $D$ is latent channel dimension, $T'=T/2^2, H'=H/2^{3}, W'=W/2^{3}$. To ensure differentiability during training, we use the reparameterization trick to sample latent vectors $z$ for motion, residual, and keyframe, respectively:
\begin{equation}
z = \mu + \sigma \odot \epsilon, \quad \epsilon \sim \mathcal{N}(0, I), \quad \sigma = \exp\left(\frac{\log\sigma^2}{2}\right)
\end{equation}
where $\epsilon$ represents a standard normal noise, and $\sigma$ is the latent standard deviation. Then the latent vectors $z_m, z_r, z_k$ are fed into a shared 3D decoder $\mathcal{D}$, which restores spatiotemporal resolution. This parameter-sharing mechanism in the decoder ensures architectural efficiency while avoiding cross-component feature interference. Specifically, each latent component is processed by the shared decoder to generate its corresponding reconstructed output:  
\begin{equation}
\hat{\mathcal{X}}_i = \mathcal{D}_i(z_i), \quad i \in \{k, m, r\}
\end{equation}  
Finally, they are fed back for video recoupling, as detailed in the next section.

\subsection{Decoupled Video Reconstruction}

This section details the end-to-end process of decoupling raw videos into components, constructing VAE inputs, and recoupling decoded components into video frames. Previous video VAE methods directly learn latent representation from raw pixel space, which fail to effectively exploit spatiotemporal information. To solve this issue, Our DeCo-VAE first decouples a video sequence into distinct components, as shown in Fig.~\ref{Overview} (b), verifying the effectiveness of removing spatial redundancy in the video sequence. Formally, given an input video frame sequence $\mathcal{X} = \{x_0, x_1, ..., x_{T-1}\}$ ($X \in \mathbb{R}^{3 \times T \times H \times W}$), we first select the first frame $x_0$ as the reference keyframe. For each frame $x_t (0\leq t \leq T-1 )$, we concatenate the current frame $x_t$ with the keyframe $x_0$ to form an input pair and use a pre-trained motion module $\mathcal{M}$ to model inter-frame geometric transformations. The module $\mathcal{M}$ outputs a motion tensor $m_t$, containing optical flow and local scale field information:
\begin{equation}
m_t = \mathcal{M}(\text{Concat}(x_t, x_0) ) \in \mathbb{R}^{3 \times H \times W}
\end{equation}
We warp the keyframe $x_0$ using $m_t$ via a differentiable warping operation $\mathcal{W}$ to generate a motion-predicted frame $\hat{x}_t^{\text{m}}$:
\begin{equation}
\hat{x}_t^{\text{m}} = \mathcal{W}(x_0, m_t) \in \mathbb{R}^{3 \times H \times W}
\end{equation}
This frame captures the global dynamic similarity between $x_t$ and $x_0$ but lacks fine-grained details (e.g., texture edges). The residual $r_t$ is defined as the pixel-wise difference between the original frame $x_t$ and $\hat{x}_t^{\text{m}}$:
\begin{equation}
r_t = x_t - \hat{x}_t^{\text{m}} \in \mathbb{R}^{3 \times H \times W}
\end{equation}
We feed motion, residual, and keyframe into dedicated encoders:
\begin{equation}
\mathcal{X}_m = \{m_t\}_{t=0}^{T-1}, \quad \mathcal{X}_r = \{r_t\}_{t=0}^{T-1}, \quad \mathcal{X}_k = T \otimes x_0
\label{eq:vae_input}
\end{equation}
VAE decodes latent representation to reconstruct motion $\hat{m}_t$ and residual $\hat{r}_t$, reconstructed keyframe $\hat{x}_0$ is obtained by averaging $\hat{\mathcal{X}}_{\text{k}}$ along the channel dimension, and recouples them to generate the final frame $\hat{x}_t$:
\begin{equation}
\hat{x}_t = \mathcal{W}(\hat{x}_0, \hat{m}_t) + \hat{r}_t
\end{equation}
Finally, we obtain the reconstructed video frames $\hat{\mathcal{X}} = \{\hat{x}_t\}_{0}^{T-1}$.
This recoupling enforces the same dynamic logic as the original decoupling process, ensuring spatiotemporal consistency of the reconstructed video.

\begin{table*}[ht]
 \centering
 \renewcommand{\arraystretch}{1.0}
 \setlength{\tabcolsep}{4pt} 
 \scalebox{0.76}{
\begin{tabular}{c| cc |cccc |cccc}
 \toprule[1.2pt]
   \multirow{2}{*}{\textbf{Method}}  & \multirow{2}{*}{\textbf{Compression Rate}} & \multirow{2}{*}{\textbf{Channels}} & \multicolumn{4}{c}{\textbf{WebVid}} & \multicolumn{4}{c}{\textbf{Kinetics-400}}\\
   \cmidrule(lr){4-11} 
   &&&  \textbf{PSNR} $(\uparrow)$ & \textbf{SSIM}$(\uparrow)$ & \textbf{LPIPS} $(\downarrow)$ & \textbf{rFVD} $(\downarrow)$ &\textbf{PSNR} $(\uparrow)$ & \textbf{SSIM}$(\uparrow)$ & \textbf{LPIPS} $(\downarrow)$ & \textbf{rFVD} $(\downarrow)$ \\
\midrule
OD-VAE (arXiv:2412)~\cite{open_sora_plan} &$4\times8\times8$ &4 &31.05	&0.8650	&0.0590 &299.60 &31.88	&0.9042	&\underline{0.0471}
	&194.00 \\

CV-VAE (NeurIPS'24)~\cite{cvvae} &$4\times8\times8$ &4 &29.71	&0.8425	&0.1295 &537.16 &29.64	&0.8736	&0.0899	&328.98 \\

WF-VAE (CVPR'25)~\cite{wfvae} &$4\times8\times8$ &16 &\underline{31.37}	&\underline{0.8961}	&\underline{0.0538} &\underline{158.90} &\textbf{34.72}	&\textbf{0.9392}	&\textbf{0.0288}	&\textbf{85.32} \\

VidTwin (CVPR'25)~\cite{vidtwin} &- &- &30.67 &0.8594	&0.1413 &593.34 &29.95	&0.8782	&0.1034	&518.80 \\

LeanVAE (ICCV'25)~\cite{leanvae} &$4\times8\times8$ &16 &29.73	&0.8615	&0.0723  &218.26 &30.86	&0.8979	&0.0543	&219.96 \\

\midrule

\rowcolor[RGB]{251,231,207}
DeCo-VAE (Ours) &$4\times 8\times 8$ &16 &\textbf{32.29} &\textbf{0.9098} &\textbf{0.0491} &\textbf{121.66} &\underline{32.30}	&\underline{0.9200}	&0.0570	&\underline{167.85} \\

\bottomrule[1.2pt]
\end{tabular}}
    \vspace{-0.4em}
 \caption{\textbf{Quantitative results of video reconstruction.} Our DeCo-VAE achieved superior performance on the WebVid~\cite{bain2021frozen} and Kinetics-400~\cite{kay2017kinetics} datasets. The first best result is highlighted in \textbf{bold}, and the second best result is \underline{underlined}. }
\label{tb:Comparison} 
\end {table*}

\begin{table}[t]
\centering
\renewcommand{\arraystretch}{1.20}
\begin{tabular}{c | c | c}
\toprule[1.2pt]
\textbf{Method} & \textbf{Channels} & \textbf{$FVD_{16} (\downarrow)$ } \\
\midrule
VideoGPT & - & 2880.6 \\
StyleGAN-V & - & 1431 \\
LVDM & - & 372 \\
Latte & - & 477.97 \\
LeanVAE-Latte & 16 & \underline{175.33} \\
WF-VAE-Latte & 16 & 371.15 \\
\midrule
DeCo-VAE-Latte (Ours) & 16 &  \textbf{166.39} \\
\bottomrule[1.2pt]
\end{tabular}
\vspace{-0.2em}
\caption{\textbf{Video generation results of different video VAEs on the UCF101~\cite{ucf101} dataset.} Our method improves the performance on downstream video generation task. }
\label{tb:ucf101_generation}
\vspace{-1.2em}
\end{table}





\subsection{Decoupled Adaptation Strategy}

To address cross-component feature entanglement and ensure stable training for DeCo-VAE, we propose a decoupled adaptation strategy, which isolates the learning of static and dynamic features through sequential phase-wise training, leveraging selective encoder freezing to avoid interference while preserving spatiotemporal consistency via the shared decoder.  

The training process is structured into two sequential phases, with the shared 3D decoder kept trainable throughout to maintain coherence across components:  

\textbf{Phase 1: Static Feature Foundation.} We freeze motion module to prevent dynamic features from disrupting static learning. During this phase, we train keyframe encoder, motion encoder, residual encoder, and shared decoder. This prioritizes learning static appearance and dynamic prior information. By pretraining motion module, we feed the decoupled components into VAE to establish a stable baseline for spatial consistency.  

\textbf{Phase 2: Dynamic Feature Refinement.} We freeze keyframe encoder to preserve pre-learned static features, then train motion module, motion encoder, residual encoder, and the shared decoder. This phase focuses exclusively on modeling inter-frame dynamics, ensuring dynamic features are learned without overwriting or entangling with static ones. This staged isolation eliminates cross-component interference, enabling precise learning of both static and dynamic characteristics.

During our training process, we employ reconstruction loss, perceptual loss, and KL regularization loss to learn basic video reconstruction capabilities. In the later training stage, we introduce a Generative Adversarial Network (GAN) and further optimize generation quality via adversarial loss. The discriminator network aids in improving the visual realism of reconstructed videos.   

The total loss integrates all loss terms to balance basic reconstruction, visual realism, and temporal consistency:  
\begin{equation}
\begin{split}
\mathcal{L}_{\text{total}} = \lambda_{\text{recon}}\mathcal{L}_{\text{recon}} + \lambda_{\text{kl}}\mathcal{L}_{\text{kl}} + \lambda_{\text{adv}}\mathcal{L}_{\text{adv}}+\lambda_{\text{p}}\mathcal{L}_{\text{p}}
\end{split}
\end{equation}
where $\mathcal{L}_{recon}$ is the reconstruction loss, $\mathcal{L}_{kl}$ is the KL divergence loss, $\mathcal{L}_{adv}$ is the adversarial loss, and $\mathcal{L}_{p}$ is the perceptual loss.

\section{Experiments}

\subsection{Experimental Details}

\paragraph{Datasets}  We trained our model on the Kinetics-400 train dataset~\cite{kay2017kinetics}, and conducted evaluations on the Webvid~\cite{bain2021frozen}  and Kinetics-400 valid datasets. To assess the performance of video VAE methods, we employed PSNR~\cite{hore2010image}, SSIM~\cite{wang2004image}, LPIPS~\cite{zhang2018unreasonable} and reconstruction FVD (rFVD)~\cite{fvd} as metrics for evaluating reconstruction quality. Kinetics-400 is a large-scale, high-quality video dataset curated from YouTube, encompassing a diverse range of human actions. It comprises 400 human action classes, with each class containing at least 400 video clips. WebVid is a large-scale text-video paired dataset, consisting of 10 million video-text pairs scraped from websites, we only use WebVid-val as our test set. To evaluate performance of video generation in diffusion model with our DeCo-VAE, we employed the UCF-101 dataset~\cite{ucf101} to train diffusion model with the base of Latte~\cite{latte}. We calculated the $FVD_{16}$ to compare the different generation results.

\paragraph{Implementation Details}  For training DeCo-VAE, all datasets were resized to 256 $\times$ 256 and the number of video frame is $17$. The training was performed on 8 NVIDIA H200-140GB GPUs, we adopted Adam~\cite{adam} optimizer with $\beta_1=0.5$ and $\beta_2=0.9$, batch size of $5$ per GPU, learning rate of $5 e^{-5}$, and total step is $500000$. KL weight $\lambda_{kl}$ was $1e^{-7}$, reconstuction loss weight $\lambda_{recon}$ and perceptual loss weight $\lambda_{p}$ was $4.0$, with the start of $400000$ steps we opened the GAN adversarial loss and the loss weight $\lambda_{adv}$ is $0.2$. We set $400000$ steps as training phase 1 with frozen motion module, and the last $100000$ steps as training phase 2 with frozen keyframe encoder.

\begin{figure}[t]
\begin{center}

\includegraphics[width=1\linewidth]{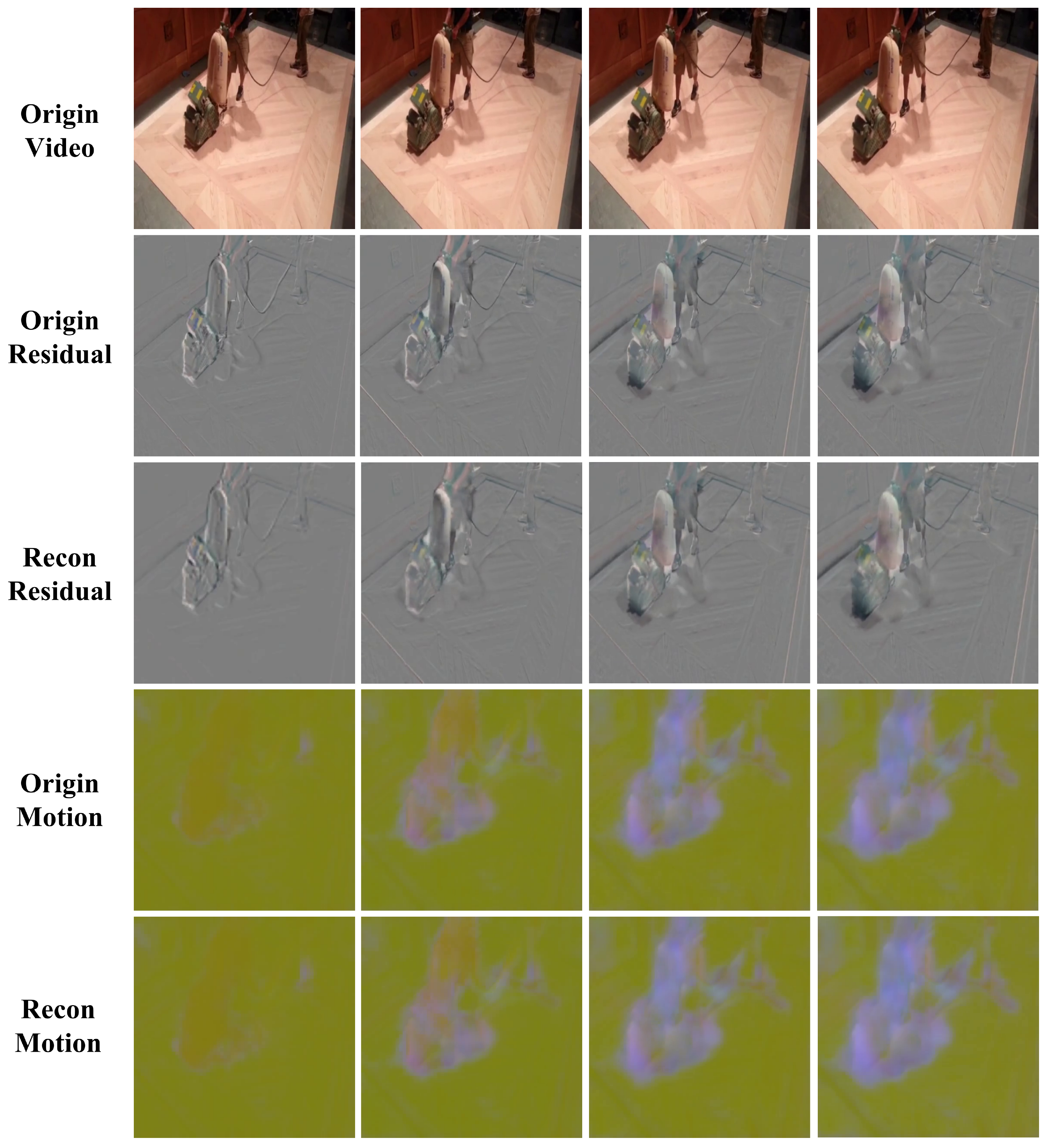}

\end{center}
\caption{\textbf{Visualization of decoupled components and their VAE reconstructions.} We showed original video frames, raw decoupled components (residual, motion), and their reconstructions by DeCo-VAE. Close alignment confirms the model’s ability to precisely reconstruct distinct decoupled features.} \label{Decouple_Vis}
\end{figure}

\begin{figure*}[t]
\begin{center}

\includegraphics[width=0.94\textwidth]{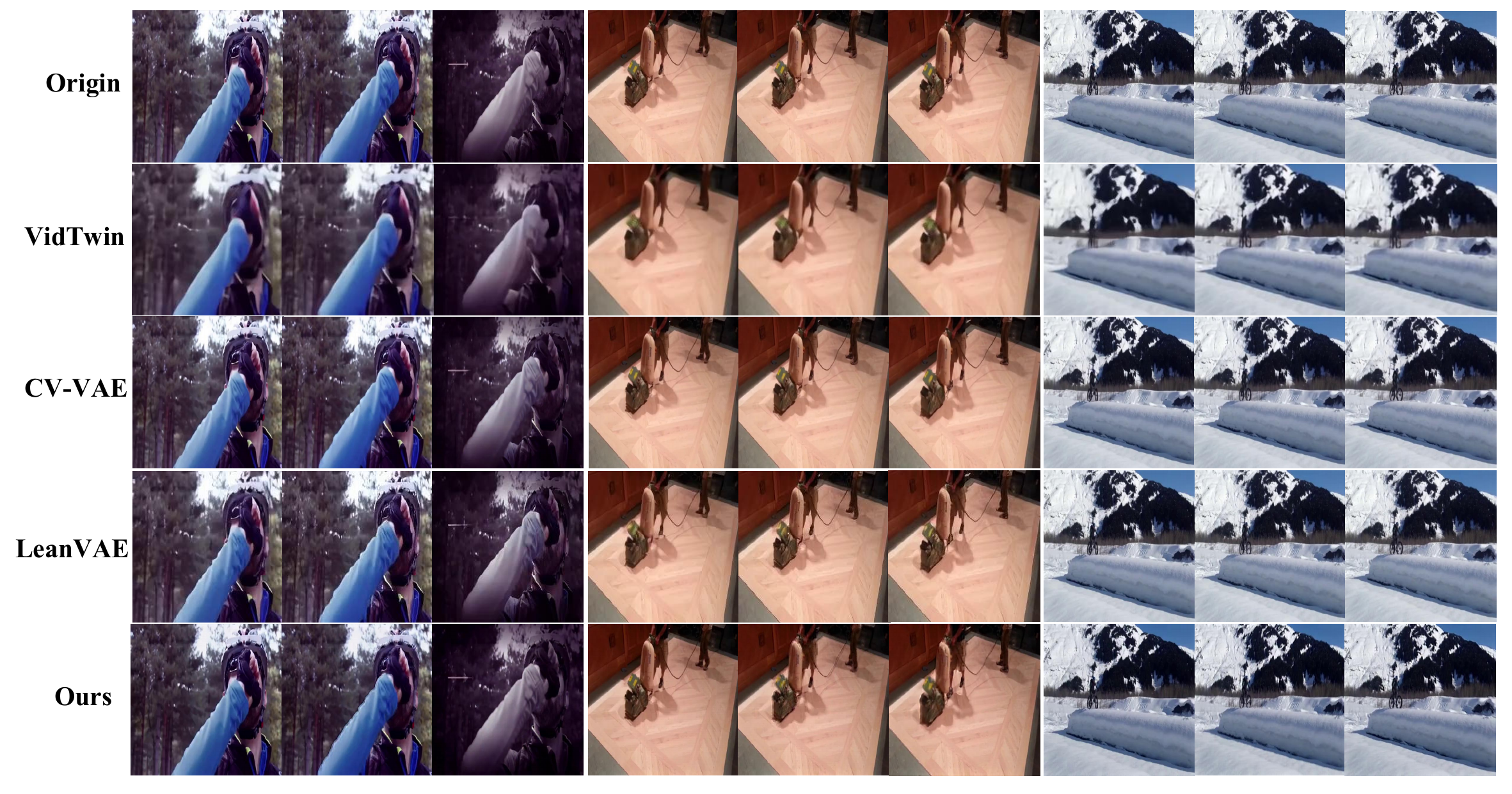}

\end{center}
\caption{\textbf{Video reconstruction results of different methods.} We compared the original video with outputs of VidTwin~\cite{vidtwin}, CV-VAE~\cite{cvvae}, LeanVAE~\cite{leanvae}, and our DeCo-VAE across three video sequences. Our method achieved superior reconstruction aligned with the original. } \label{Comparison}
\end{figure*}

\subsection{Comparison with SoTA methods}

We compared our DeCo-VAE to other SoTA methods, including OD-VAE~\cite{open_sora_plan}, CV-VAE~\cite{cvvae}, WF-VAE~\cite{wfvae}, VidTwin~\cite{vidtwin}, LeanVAE~\cite{leanvae}. Following previous work, we reported video reconstruction quality on $256 \times 256 \times 17$ video clips.

\paragraph{Quantitative Evaluation}
The comparison results were illustrated in Tab.~\ref{tb:Comparison}, while the parameters comparison are shown as Tab.~\ref{tb:model_params}. All compared methods (except VidTwin with unspecified compression rate) adopted the same $4 \times 8 \times 8$ compression setting, our DeCo-VAE achieved superior overall performance while maintaining lightweight parameters. On the WebVid dataset, DeCo-VAE outperformed all baselines across all metrics (PSNR, SSIM, LPIPS, rFVD), obtaining the best results. On the Kinetics-400 valid dataset, it attained the second-best results in PSNR, SSIM and rFVD, with competitive LPIPS performance. In summary, DeCo-VAE achieved SoTA reconstruction quality under the $4 \times 8 \times 8$ compression setting with lightweight network, providing an efficient latent representation for downstream generative tasks.

\paragraph{Qualitative Evaluation} Fig.~\ref{Comparison} visually compares frames reconstructed by DeCo-VAE and three representative baselines (WF-VAE, CV-VAE, OD-VAE) under $4 \times 8 \times 8$ compression. DeCo-VAE restored fine textures and motion boundaries more faithfully, while maintained the static regions. We validated DeCo-VAE reconstruction capability of the decoupled components via Fig.~\ref{Decouple_Vis}, which visualized the original residual and motion components, alongside their reconstructions. A closer look at the visualization reveals striking fidelity in both component types. The tight alignment between original and reconstructed components directly demonstrated that DeCo-VAE’s explicit decoupling design enables precise capture and recovery of semantically distinct features, laying a foundation for high-quality video reconstruction.

\paragraph{Generation Performance} To evaluate the effectiveness of our proposed DeCo-VAE architecture in enhancing video generation capabilities, we integrated it into the Latte model and conducted comprehensive comparative experiments against a series of state-of-the-art video VAE methods. Detailed comparison results are summarized in Tab.~\ref{tb:ucf101_generation}, where the $FVD_{16}$ serves as the core evaluation criterion. Notably, the diffusion model equipped with our DeCo-VAE achieved a superior $FVD_{16}$ score of 166.39 when using 16 channels, which outperforms all existing methods. This clearly demonstrated that our DeCo-VAE method improves the overall performance of the downstream video generation task.

\begin{table}[t]
\vspace{1.2em}
\centering
\renewcommand{\arraystretch}{1.20}
\begin{tabular}{c | c | c | c}
\toprule[1.2pt]
\textbf{Model} & \textbf{Channels} & \textbf{rFVD} ($\downarrow$) & \textbf{Param.} ($\downarrow$) \\ 
\midrule
OD-VAE & 4 & 299.60 & 239M \\
CV-VAE & 4 & 537.16 & 182M \\ 
WF-VAE & 16 & \underline{158.90} & 316M \\
VidTwin & - & 593.34 & 157M \\
LeanVAE & 16 & 218.26 & \textbf{40M} \\
\midrule
DeCo-VAE (Ours) & 16 & \textbf{121.66} & \underline{62M} \\
\bottomrule[1.2pt]
\end{tabular}
\vspace{-0.2em}
\caption{\textbf{Comparison of model parameters across different methods.} The first best result is highlighted in \textbf{bold}, and the second best result is \underline{underlined}. }
\label{tb:model_params}
\vspace{-1.2em}
\end{table}

\begin{figure}[t]
\begin{center}

\includegraphics[width=1\linewidth]{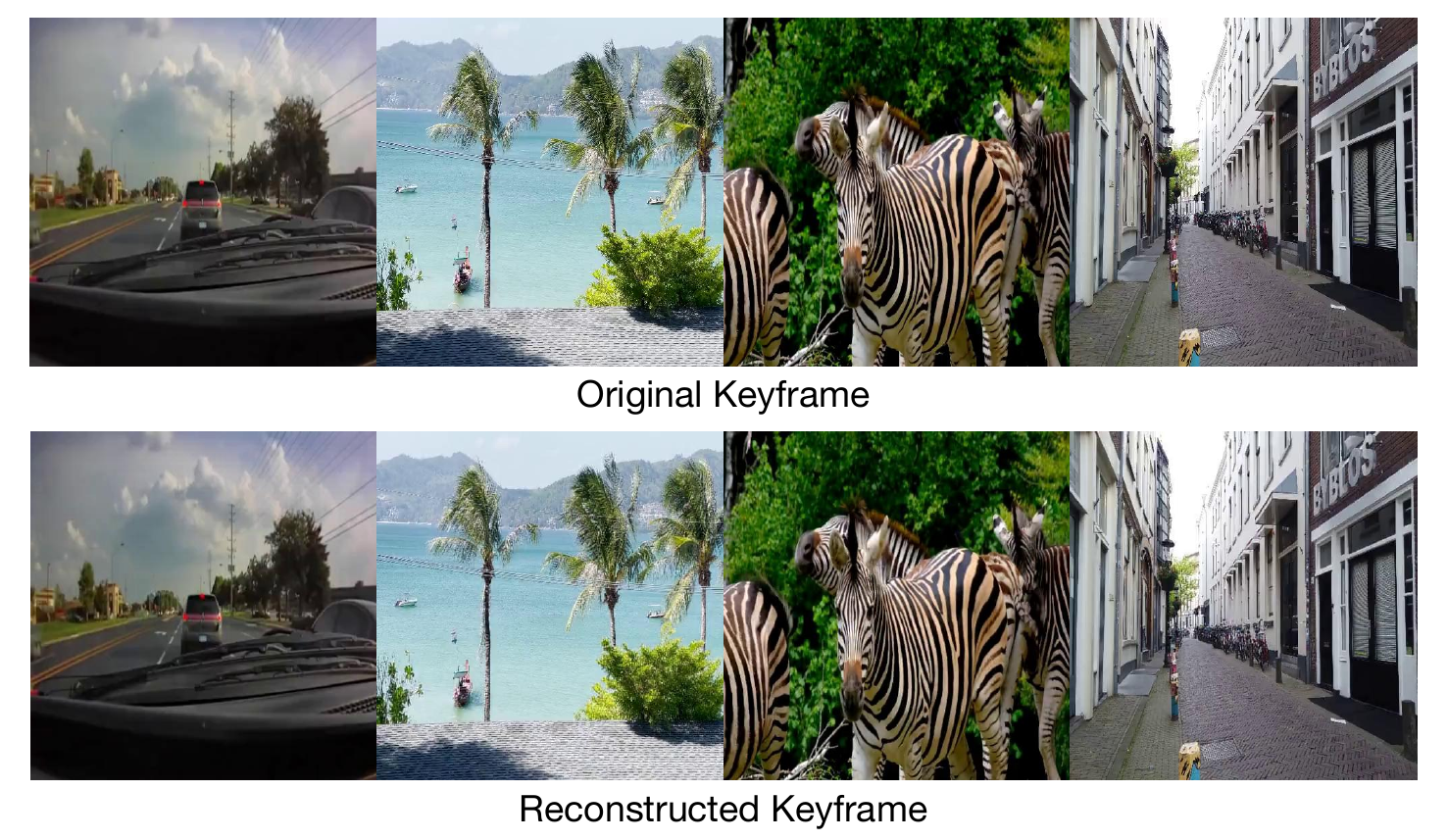}

\end{center}
\caption{\textbf{Visualization of keyframes and their VAE reconstructions.} Our keyframe encoder achieved good latent representation and could reconstruct clear keyframe to recouple the components.} \label{Keyframe_Vis}
\end{figure}

\subsection{Ablation Studies}

\paragraph{Model Architecture} We conducted ablation studies to verify the effectiveness of core components: video decoupling (V.D.) with dedicated encoders and decoupled adaptation strategy (D.A.). The results are shown in Tab.~\ref{tb:ablation}. When both components were disabled (top row), the model achieved 29.80 dB PSNR, 0.8640 SSIM, and 0.0718 LPIPS. Enabling only V.D. (middle row) significantly improved performance: PSNR rises by 1.40 dB to 31.20, and SSIM increases by 0.0289 to 0.8929, confirming that video decoupling effectively preserves key details by separating encoding branches. Further enabling the D.A. strategy (bottom row) brings additional gains: PSNR climbs to 32.29 dB (a further 1.09 dB increase), SSIM reaches 0.9098, and LPIPS drops sharply to 0.0491 (a 33.8\% reduction compared to the middle row). This validated that branch-specific fine-tuning suppresses cross-talk between motion and residual branches, optimizing overall reconstruction quality. We visualized the keyframe encoder of DeCo-VAE, as shown in Fig.~\ref{Keyframe_Vis}. The results demonstrated that our keyframe encoder learned more compact latent representation of keyframes, while the shared 3D decoder reconstructed clear keyframes and enhances the recoupling process, this verified the effectiveness of our dedicated encoders.

We compared network designs for decoupled components as shown as Tab.~\ref{tb:ablation_encoder}. Directly concatenating keyframes, motion, and residuals along the channel dimension enabled the PSNR and SSIM reducing obviously to 27.15 and 0.8081, respectively. This demonstrated that the mixing of decoupled components leads to latent representation conflicts, while our design of dedicated encoders enabled full use of the advantages of video decoupling to achieve compact latent space.


\begin{table}[t]
\centering
\renewcommand{\arraystretch}{1.20}
\setlength{\tabcolsep}{4pt} 
\scalebox{0.97}{
\begin{tabular}{c c c c}
\toprule[1.0pt]
\multirow{2}{*}{\textbf{Settings}} & \multicolumn{3}{c}{\textbf{WebVid}} \\
\cmidrule(lr){2-4}
  & \textbf{PSNR} $(\uparrow)$ & \textbf{SSIM} $(\uparrow)$ & \textbf{LPIPS} $(\downarrow)$ \\
\midrule
 
Concat & 27.15	&0.8081	&0.1379 \\

Dedicated Encoders & 31.20	&0.8929	&0.0741 \\

\bottomrule[1.2pt]
\end{tabular}}
\vspace{-0.2em}
 \caption{\textbf{Ablation studies on decoupled design.} "Concat" refers to directly concatenating keyframes, motion, and residual along the channel dimension, which are then fed into a single VAE encoder. "Dedicated Encoders" refers to employing distinct encoders for each of the decoupled components. }
 \label{tb:ablation_encoder}
\end{table}

\begin{table}[t]
\centering
\renewcommand{\arraystretch}{1.20}
\scalebox{0.97}{
\begin{tabular}{c c c c c}
\toprule[1.2pt]
\multicolumn{2}{c}{\textbf{Settings}} & \multicolumn{3}{c}{\textbf{WebVid}} \\
\cmidrule(lr){1-2}\cmidrule(lr){3-5}
\textbf{V. D.} & \textbf{D. A.} & \textbf{PSNR} $(\uparrow)$ & \textbf{SSIM} $(\uparrow)$ & \textbf{LPIPS} $(\downarrow)$ \\
\midrule
 &  &29.80	&0.8640	&0.0718 \\
\ding{51} &  &31.20	&0.8929	&0.0741 \\

\ding{51} & \ding{51} &32.29	&0.9098	&0.0491 \\

\bottomrule[1.2pt]
\end{tabular}}
\vspace{-0.2em}
 \caption{\textbf{Ablation studies on model architecture.} "V. D." represents video decoupling with dedicated encoders, and "D. A." represents decoupled adaptation strategy. }
 \label{tb:ablation}
 \vspace{-1.2em}
\end{table}

\section{Conclusion}

We present DeCo-VAE, a decoupled video VAE framework that explicitly separates video content into keyframe, motion, and residual components to achieve compact and interpretable latent representation. Introducing dedicated encoders and a shared 3D decoder, DeCo-VAE effectively avoids cross-component interference while maintaining spatiotemporal coherence. A decoupled adaptation strategy further stabilizes training and enables precise learning of static and dynamic features. Extensive experiments validate that DeCo-VAE achieves outstanding reconstruction quality with lightweight design, getting superior results on the WebVid and Kinetics-400 datasets with PSNR, SSIM and LPIPS. This provides efficient and versatile latent representation for downstream video generation and modeling tasks, which also excels in low-resource deployment scenarios and supports seamless integration with various task-specific fine-tuning pipelines.

\section{Limitations and Future Work}

DeCo-VAE excels at short videos but struggles with longer sequences due to single-keyframe reliance. In longer clips such as scene or viewpoint shifts, the keyframe quickly becomes irrelevant, forcing motion/residual components to encode complex differences against an outdated anchor, causing bloated representation and poorer reconstruction. Keyframe errors also propagate through subsequent frames, as all derive from this sole reference.

To address these issues, future work will explore multi-keyframe decoupling. This reduces single-anchor dependence by using nearby, contextually relevant keyframes to simplify long-sequence representation. We will also mitigate error propagation by refining subsequent frames via local temporal consistency, lessening initial keyframe flaws’ impact. These tweaks will extend DeCo-VAE’s robustness to longer, dynamic videos while preserving efficiency.

{
    \small
    \bibliographystyle{ieeenat_fullname}
    \bibliography{main}
}


\end{document}